\documentclass[conference]{IEEEtran}
\usepackage{graphicx}
\usepackage{makecell}
\usepackage{threeparttable}

\usepackage{tikz}
\usepackage{tabularx}
\usepackage[table]{xcolor}
\usepackage{tcolorbox}

\newtcolorbox[auto counter]{pabox}[2][]{
title=Box~\thetcbcounter: #2,#1}
\usepackage{subcaption}
\usepackage{amssymb} 
\usepackage{amsmath}
\usepackage{listings}
\usepackage{fancyhdr}

\usepackage[nolist]{acronym}
\begin{acronym}[MACHU]
  \acro{ai}[AI]{Artificial Intelligence}
  \acro{mdp}[MDP]{Markov decision process}
  \acro{soa}[SOA]{Service-oriented Architecture}
  \acro{toh}[ToH]{Towers of Hanoi}
  \acro{rl}[RL]{Reinforcement Learning}
  \acro{fbrl}[FBRL]{Forward-backward reinforcement learning}
  \acro{ncm}[NCM]{Neural Cognitive Modelling}
  \acro{lnn}[LNN]{Logical Neural Networks}
  \acro{llm}[LLM]{Large Language Model}
  \acro{pddl}[PDDL]{Planning Domain Definition Language}
  \acro{fcn}[FCN]{Fully-connected Network}
  \acro{ddqn}[DDQN]{Double Deep Q networks}
  \acro{lif}[LIF]{Leaky Integrate-and-Fire}
  \acro{vsa}[VSA]{Vector Symbolic Architecture}
  \acro{ml}[ML]{Machine Learning}
  \acro{mbrl}[MBRL]{Relational Model-based Reinforcement Learning}
  \acro{ilp}[ILP]{Inductive Logic Programming}
  \acro{tos}[ToS] {Thought-of-Search}
  \acro{datos}[DA-ToS] {Double-Agent Thought-of-Search}
  \acro{autotos}[AutoToS] {Automated Thought-of-Search}
  
  \acro{strips}[STRIPS]{Stanford Research Institute Problem Solver}
\end{acronym}

\fancypagestyle{arxivfirst}{%
  \fancyhf{}%
  \fancyfoot[C]{%
    \scriptsize
    \fbox{\parbox{0.88\textwidth}{
      This work has been submitted to the IEEE for possible publication.
      Copyright may be transferred without notice, after which this
      version may no longer be accessible.}}%
  }%
}

\ifCLASSINFOpdf
\else
\fi

\usepackage{multirow}
\usepackage[table]{xcolor}
\usepackage{multirow}

\definecolor{pastelblue}{HTML}{D1E9FF} % High (Light Pastel Blue)
\definecolor{pastelyellow}{HTML}{FFF9C4} % Medium
\definecolor{pastelgreen}{HTML}{E8F5E9}  % Low

\begin{document}

\title{A 3D Characterization Framework for Intelligent Sequential Decision Making}

\author{
    \IEEEauthorblockN{
        Sadig Gojayev, Carolina Fortuna
    }
    \IEEEauthorblockA{
        \textit{Department of Communication Systems}, \textit{Jožef Stefan Institute}, SI-1000 Ljubljana, Slovenia \\
        \{sadig.gojayev, carolina.fortuna\}@ijs.si}

}

\maketitle

\thispagestyle{arxivfirst}
% \enlargethispage{-2\baselineskip}

\begin{abstract}
Puzzles are widely used to evaluate the reasoning capabilities of artificial intelligence (AI) systems for sequential decision making, yet approaches originating from different paradigms are rarely compared under unified conditions. To address this gap, we introduce a three-dimensional characterization framework that enables the analysts of AI methods by 1) projecting them to the Markov decision process (MDP) sequential decision making formalism, 2) degree of autonomy through human prior ranking of  their designs and, 3) skill and computational cost. Using this framework, we analyze how representative graph-based, reinforcement learning, and large language model (LLM)-based approaches differ in their design choices and performance characteristics, instantiated respectively by 
Neurosolver, forward–backward reinforcement learning (FBRL), and automated thought-of-search (AutoToS), including a double-agent extension of thought-of-search (DA-ToS). The analysis relies on the Tower of Hanoi puzzle that provides a controlled benchmark with well-defined rules and scalable complexity, enabling consistent comparison across increasing problem sizes. The 3D characterization reveals that LLM-based methods, due to their weakly constrained action-space design, shift complexity from architecture to inference-time verification, leading to substantially higher memory and runtime costs than Neurosolver and FBRL.
\end{abstract}


% \fi
\IEEEpeerreviewmaketitle

\section{Introduction} \label{sec:intro}

% What is important
The rise of agentic artificial intelligence (AI) \cite{murugesan2025rise} is expected to significantly impact the economy and society as a whole. While agents relying on AI techniques have been around for several decades \cite{castelfranchi1998modelling}, the new generations of such agents are designed around the recent breakthrough in auto-regressive models, namely large language models (LLMs) \cite{brown2020language}. These models, based on transformer architectures \cite{vaswani2017attention}, essentially learn a probabilistic language model that, conditioned by an input generates an output. These models exhibit broad coverage over human-generated text because they are trained on massive web-scale corpora. Whether the training process enables them to go beyond memorization towards eliciting abstraction and generalization that would subsequently enable reasoning is still under investigation \cite{wang2025generalization}. To improve reasoning, various training and prompting approaches such as reinforcement learning \cite{guo2025deepseek} and chain of thought \cite{wei2022chain} were proposed, however their reasoning capabilities are mostly  unclear \cite{shojaee2025illusion}.

Reasoning capabilities of models are often evaluated on puzzles \cite{estermann2024puzzles,hawkins2025enigme, shojaee2025illusion}. Some works focus more on logic puzzles \cite{estermann2024puzzles} while others focus on text-based puzzles, classified into numeric, sequence, and physics categories \cite{hawkins2025enigme}. \ac{toh} is one recently used puzzle to investigate reasoning differences between LLMs and LRMs \cite{shojaee2025illusion}. \ac{toh} is also widely used in education to develop algorithmic thinking through inductive and recursive approaches \cite{merrotsy2015tower,anderson1992induction}. It also has wide applicability beyond education, being used in logistics and robotics, where weighted variants have been proposed to determine optimal move sequences with minimal cost \cite{aguilar2016application, mehiri2024weighted, havur2013case}.

\begin{figure}[t]
    \centering
    \scriptsize % Smaller font for labels
    
    % Initial State
    \begin{subfigure}[b]{0.31\textwidth}
        \centering
        \begin{tikzpicture}[scale=0.45] % Reduced scale
            % Draw the pegs
            \draw[thick] (0,0) -- (0,1.5); % n1
            \draw[thick] (3,0) -- (3,1.5); % n2
            \draw[thick] (6,0) -- (6,1.5); % n3

            % Draw the base
            \draw[thick] (-1,0) -- (7,0);

            % Draw the disks on n1
            \draw[fill=red!70] (0-1.5,0) rectangle (0+1.5,0.5); % d1
            \draw[fill=blue!70] (0-1.0,0.5) rectangle (0+1.0,1.0); % d2

            % Label the pegs
            \node at (0,-0.6) {$1$};
            \node at (3,-0.6) {$2$};
            \node at (6,-0.6) {$3$};

            % Label the disks
            \node at (0,0.25) {$n_1$};
            \node at (0,0.75) {$n_2$};
        \end{tikzpicture}
        \caption{Initial state: $n_1, n_2$ on peg $1$}
    \end{subfigure}
    \\
    % Action State
    \begin{subfigure}[b]{0.31\textwidth}
        \centering
        \begin{tikzpicture}[scale=0.45]
            % Draw the pegs
            \draw[thick] (0,0) -- (0,1.5); 
            \draw[thick] (3,0) -- (3,1.5); 
            \draw[thick] (6,0) -- (6,1.5); 

            % Draw the base
            \draw[thick] (-1,0) -- (7,0);

            % Draw the disks
            \draw[fill=red!70] (0-1.5,0) rectangle (0+1.5,0.5); % d1
            \draw[fill=blue!70] (3-1.0,0) rectangle (3+1.0,0.5); % d2

            % Label the pegs
            \node at (0,-0.6) {$1$};
            \node at (3,-0.6) {$2$};
            \node at (6,-0.6) {$3$};

            % Label the disks
            \node at (0,0.25) {$n_1$};
            \node at (3,0.25) {$n_2$};
        \end{tikzpicture}
        \caption{Action: Move $n_2$ peg $2$}
    \end{subfigure}
    \\
    % Goal State
    \begin{subfigure}[b]{0.31\textwidth}
        \centering
        \begin{tikzpicture}[scale=0.45]
            % Draw the pegs
            \draw[thick] (0,0) -- (0,1.5); 
            \draw[thick] (3,0) -- (3,1.5); 
            \draw[thick] (6,0) -- (6,1.5); 

            % Draw the base
            \draw[thick] (-1,0) -- (7,0);

            % Draw both disks on n3
            \draw[fill=red!70] (6-1.5,0) rectangle (6+1.5,0.5); % d1
            \draw[fill=blue!70] (6-1.0,0.5) rectangle (6+1.0,1.0); % d2

            % Label the pegs
            \node at (0,-0.6) {$1$};
            \node at (3,-0.6) {$2$};
            \node at (6,-0.6) {$3$};

            % Label the disks
            \node at (6,0.25) {$n_1$};
            \node at (6,0.75) {$n_2$};
        \end{tikzpicture}
        \caption{Goal: $n_1, n_2$ on peg $3$}
    \end{subfigure}

    \caption{\ac{toh} for $n=2$ disks and $p=3$ pegs.}
    \label{fig:TOH}
\end{figure}
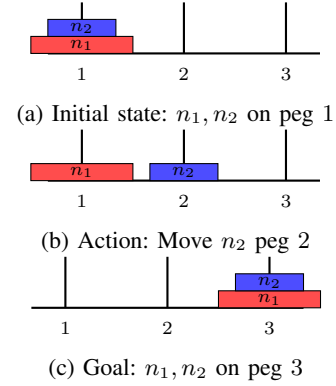

\ac{toh} involves discs $n$ and pegs $p$, where all the discs are initially placed on the first peg, and the goal is to move all the discs to the last peg. The key limitation is that only a smaller disc can be placed on top of a larger disc, as illustrated in Figure~\ref{fig:TOH}. To solve this puzzle, at each step, a human or AI player needs to understand the current state and take an action that will transition the puzzle to the next state until it is solved. This sequential decision making is referred to as a \ac{mdp} and some AI approaches, especially the reinforcement learning ones, explicitly leverage the \ac{mdp} formalism. Over time, several AI techniques, other than LLMs and LRMs, have been used to solve the ToH. For instance, \cite{edwards2018forward}  employed reinforcement learning, \cite{agravante2023learning} approached it with logical neural networks, \cite{bieszczad2015neurosolver}  proposed graph-based solvers, while \cite{stewart2011neural} investigated neural cognitive modeling.

Currently, it remains unclear what the design trade-offs are when selecting one class of AI models over another for scalable sequential decision making problems such as \ac{toh}. State-of-the-art studies mostly focus on decision-quality comparisons between LLMs and RL \cite{wang2025comparing}, LLMs and AI planners \cite{qin2024uno}. Beyond performance metrics, very recent work \cite{moerland2022unifying} provides theoretical formulations that cast LLM- and AI planners within \ac{mdp} for optimization. However, while these efforts highlight differences across paradigms, they largely lack an explicit linkage between formulation choices, decision-making skill, computational cost, and the level of human-imposed biases.

To fully characterize an AI agent, one must consider three key aspects:
(1) how the agent represents the world,
(2) the degree of external guidance or supervision it receives, and
(3) the computational cost required.
In this paper, we propose a three-dimensional characterization framework that provides a structured and systematic basis for comparing different AI methods.
 The framework captures \ac{mdp} formulation projecting to understand each method’s approach to sequential decision making, human prior to provide clearer insight into required guidance, and skill and computational cost to illustrate resource requirements. To illustrate the applicability of the proposed framework we employ it for comparing selected AI techniques.  Our key contributions are as follows:

\begin{itemize}

    \item We introduce a 3D characterization framework for sequential decision making methods that jointly analyzes (i) technique projection to the \ac{mdp}-formalism, (ii) human prior across architecture/training/inference, and (iii) skill–cost trade-offs.

    \item We propose an operational human-prior rubric with explicit levels (High/Medium/Low) for representation/mechanism, knowledge/learning, and guidance/decision/validation, enabling qualitative ranking across paradigms.
    
    \item We instantiate the framework on \ac{toh} by comparing a graph-based solver (Neurosolver)  \cite{bieszczad2015neurosolver}, a RL method \ac{fbrl} \cite{edwards2018forward}, and LLM-based planning via \ac{autotos} \cite{katz2024thought}, including a new \ac{datos} variant that modularizes code generation and repair. 

    \item We show that weakly constrained LLM action spaces shift complexity from model design to inference-time verification loops, resulting in substantially higher runtime/memory than graph-based/RL approaches under the same task scaling.

\end{itemize}

This paper is organized as follows. Section~\ref{sec:rel_work} summarizes related work,  Section~\ref{sec:3D} introduces the three dimensional characterization framework while Section~\ref{sec:methods} elaborates on the  methods. Sections~\ref{sec:dim1}, \ref{sec:humanprior} and \ref{sec:results} analyze the selected AI methods from the perspective of the proposed framework. Finally, Section~\ref{sec:con} concludes the paper.

\section{Related Work}\label{sec:rel_work}

Comparisons of sequential decision making methods across different paradigms and backgrounds usually focus on outcome-based metrics, such as success rate or optimality. The UNO Arena \cite{qin2024uno} benchmark evaluates agents from reinforcement learning, heuristic, and LLM-based approaches by emphasizing decision-level quality, using metrics such as hit rate and decision rank rather than only win rate. For puzzle-solving tasks, PUZZLES \cite{estermann2024puzzles} provides a benchmark of logical puzzles for neural reasoning in reinforcement learning, evaluating various RL algorithms primarily through success rate. PUZZLEWORLD \cite{li2025puzzleworld}, in contrast, evaluates open- and closed-source LLMs by focusing on stepwise accuracy, where models are tasked with producing step-by-step reasoning toward a final solution.
Recent work by Wang et al. \cite{wang2025comparing} compares human decision-making with LLM and reinforcement learning agents, evaluating coordination behavior and path cost under identical environmental conditions. Overall, these comparisons adopt a largely single computational perspective, focusing primarily on result-based measures such as inference quality, even across different paradigms. They emphasize external performance metrics while lacking explicit analysis and direct linkage to representation-level action spaces, architecture-level transition mechanisms, and human guidance, leaving internal decision-making processes treated as black boxes.

Beyond purely computational evaluations, theoretical efforts such as FRAP \cite{moerland2022unifying} decompose AI planning and reinforcement learning algorithms into explicit dimensions (e.g., root states, trials, bootstrapping, backups) and analyze them through the lens of \ac{mdp} optimization. Similarly, Burden et al. \cite{burden2025paradigms} classify AI evaluation approaches into paradigms such as benchmarking and exploratory evaluation. While these frameworks provide valuable formulations for comparing methods across paradigms, they do not offer insight into decision-making skill and computational cost in relation to \ac{mdp} formulation. As a result, it remains unclear how action-space design, human guidance, and architectural constraints affect inference quality and computational cost.

There is a lack of unified frameworks across different paradigms, particularly between reinforcement learning and large language model–based approaches. Most comparisons focus on outcome metrics, rather than the formulation of sequential decision making (e.g., \ac{mdp}s) and the role of human insight in shaping model behavior.

\section{Proposed 3D Characterization Framework}
\label{sec:3D}
In this section, we introduce a three-dimensional framework for comparing AI methods used in sequential decision making. The framework focuses on three aspects: how each method projects to the elements of a \ac{mdp}, what is the level of human prior introduced  throughout the AI model development, and what skill vs computational trade-offs it involves. 

\subsection{Projecting through Markov Decision Process}
The first dimension, projecting to \ac{mdp}, clarifies how each method handles the core components of sequential decision making—such as state representation, transitions, rewards, and goal identification. Understanding this projection shows how a method reasons about progress and how it identifies correct or optimal next steps. Sequential decision making is typically formulated as a \ac{mdp}~\cite{chen2024ai}, which is defined as $\mathcal{M} = \langle S, A, T, R, G \rangle$, where states ($S$) denote environment configurations, actions ($A$) specify possible moves, the transition function ($T$) gives the probability of reaching $s'$ after taking action $a$ in state $s$, the reward function ($R$) assigns values to transitions, and goal states ($G$) indicate task completion.  

\begin{table}[h!]
\begin{tabularx}{\linewidth}{>{\centering\arraybackslash}p{0.8cm} X}
\hline

\rowcolor{gray!10}
\multicolumn{2}{p{0.95\linewidth}}{
\textbf{State space representation} refers to how the \ac{mdp} state space $S$ is specified in the sense of \ac{mdp}. The levels of human prior are defined as follows:
} \\
\hline
\textbf{High} &
Humans explicitly design the state representation, fully specifying its structure and semantics, as in symbolic planners based on PDDL \cite{silver2022pddl}. \\
\textbf{Medium} &
Humans prescribe a high-level representational structure or template, while allowing the system to learn internal parameters or details, as in logical neural networks or partially structured encodings \cite{agravante2023learning}. \\
\textbf{Low} &
No explicit representational structure is imposed by humans; the model learns state representations end-to-end from interaction or data, as in fully end-to-end reinforcement learning systems trained via self-play \cite{bai2020provable}. \\
\hline
\hline
\rowcolor{gray!10}
\multicolumn{2}{p{0.95\linewidth}}{
\textbf{Mechanism} refers to the algorithmic structure responsible for reasoning and inference. The levels of human prior are defined as follows:
} \\
\hline
\textbf{High} &
Humans design the inference procedure, such as logical rules or search heuristics. \\
\textbf{Medium} &
Humans select the general type of mechanism to employ, while the algorithm handles the details; for example, in Value Iteration Networks \cite{tamar2016value}, humans prescribe a value-iteration–based planning structure, while the network learns the reward and transition representations end-to-end. \\
\textbf{Low} &
Humans provide only high-level instructions, such as requesting an LLM to ``generate code for running a DDQN,'' without specifying internal structure. \\
\hline
\end{tabularx}
\caption{Levels of Human Prior in Architecture Design.}
\label{tab:architecture}
\end{table}

\begin{table}
\begin{tabularx}{\linewidth}{>{\centering\arraybackslash}p{0.8cm} X}
\hline

\rowcolor{gray!10}
\multicolumn{2}{p{0.95\linewidth}}{
\textbf{Knowledge Input} concerns how training data, constraints, or reward structures are introduced, directly shaping the transition function $T$ and reward function $R$ in \ac{mdp} formulation.
} \\
\hline
\textbf{High} &
Humans explicitly specify the training data, define relevant constraints, and design or shape reward structures, fully guiding how the model learns from examples, as in reinforcement learning with human-structured environments and hand-crafted training scenarios (e.g., Q-learning with expert-designed tasks). \\
\textbf{Medium} &
Humans provide curated or loosely structured data without fully specified objectives, allowing the system to organize patterns and infer useful structure while still retaining partial human guidance, such as transformers trained on human-curated datasets. \\
\textbf{Low} &
The model autonomously extracts training signals from large-scale interaction or self-generated experience, with minimal human specification beyond initial setup, as in self-supervised pretraining of large language models on unlabeled data \cite{hegde2024self}. \\
\hline
\hline
\rowcolor{gray!10}
\multicolumn{2}{p{0.95\linewidth}}{
\textbf{Learning} reflects how the model updates its parameters in response to the training signal and the extent of human involvement in this process.
} \\
\hline
\textbf{High} &
Humans explicitly shape the optimization process through curated objectives, staged training procedures, or tightly controlled learning phases, as in reinforcement learning from human feedback (RLHF) for large language models. \\
\textbf{Medium} &
Humans specify general optimization settings—such as learning rates, network architectures, or update rules—while the system autonomously manages the internal learning dynamics, as in Double Deep Q-Networks (DDQN) \cite{van2016deep}. \\
\textbf{Low} &
Learning proceeds end-to-end with minimal human involvement beyond initialization, with the system autonomously generating training signals, as in self-play reinforcement learning \cite{bai2020provable}. \\
\hline
\end{tabularx}
\caption{Levels of Human Prior at Training Stage.}
\label{tab:training}
\end{table}

\begin{table}[h!]
\begin{tabularx}{\linewidth}{>{\centering\arraybackslash}p{0.8cm} X}
\hline

\rowcolor{gray!10}
\multicolumn{2}{p{0.95\linewidth}}{
\textbf{Guidance} captures the extent to which humans specify task details or constrain model behavior at runtime.
} \\
\hline
\textbf{High} &
Humans provide detailed task instructions, including explicit constraints, edge cases, and step-level guidance, often through fully specified prompts or rule descriptions, as in Least-to-Most prompting techniques \cite{zhou2022least}. \\
\textbf{Medium} &
Humans specify high-level task constraints or goals without defining the complete execution structure, such as providing domain rules (e.g., ``a smaller disk cannot be placed beneath a larger one'' in \ac{toh}) while leaving reasoning to the model. \\
\textbf{Low} &
Humans provide only minimal task information—such as specifying the start and goal states in graph-based solvers—while the system autonomously determines the reasoning process and action sequence. \\
\hline
\hline
\rowcolor{gray!10}
\multicolumn{2}{p{0.95\linewidth}}{
\textbf{Decision} describes how actions or outputs are selected by the system during inference and the extent of human involvement in this process.
} \\
\hline
\textbf{High} &
Humans explicitly drive or supervise the decision-making process, often through interactive or step-wise instructions. The model depends on human-provided reasoning steps, clarifications, or corrective feedback to determine each action, as in interactive planning with large language models (e.g., DEPS) \cite{wang2023describe}. \\
\textbf{Medium} &
Humans specify the decision framework or evaluation criteria—such as accuracy thresholds, selection rules, or preferred heuristics—while the system autonomously executes decisions within these constraints, as in safe reinforcement learning approach in constrained \ac{mdp} \cite{wachi2020safe}. \\
\textbf{Low} &
The system selects actions entirely autonomously based on its learned internal dynamics or probability distributions, without any human-defined decision logic at inference time, as in end-to-end transformer-based language models. \\
\hline
\hline
\rowcolor{gray!10}
\multicolumn{2}{p{0.95\linewidth}}{
\textbf{Validation} describes how the correctness of model outputs is assessed and the extent of human involvement in this process.
} \\
\hline
\textbf{High} &
Humans manually inspect model outputs, identify errors, and provide corrective feedback after each execution. \\
\textbf{Medium} &
Humans specify validation criteria or automated test suites, while the system autonomously executes these checks and uses the results to guide refinement or improving tests, as in TestGen-LLM framework \cite{alshahwan2024automated}. \\
\textbf{Low} &
Validation and correction are performed entirely by the system itself, without human oversight, as in self-correcting large language models \cite{welleckgenerating}. \\
\hline
\end{tabularx}
\caption{Levels of Human Prior at Inference Stage.}
\label{tab:inference}
\end{table}

\subsection{Human Prior in AI for Sequential Decision Making} \label{sec:human_prior}
The second dimension, human prior, reflects how much human know-how is embedded in a method  during its development: architecture design, training, and inference stages. This helps assess the degree of intrinsic intelligence that characterizes a method. We propose a  categorization of human prior in AI techniques for sequential decision making focusing on three major aspects: architecture design, model training and model inference as follows.
\paragraph{Architecture Design}
The architecture design involves the development of the structure of the model prior to learning or execution. We identify two components in the design: \textit{Representation} and \textit{Mechanism} as defined in Table \ref{tab:architecture}.

\paragraph{Training Stage}
The training stage describes how a model acquires skill and how humans shape this process. It consists of two substages: \textit{Knowledge Input}  and \textit{Learning} as defined in Table \ref{tab:training}.

\paragraph{Inference Stage}
Inference describes how the model behaves once training is complete. It includes how humans guide the model, how decisions are produced, and how outputs are checked. It consists of three substages: \textit{Guidance}, \textit{Decision}  and \textit{Validation} as defined in Table \ref{tab:inference}.

\begin{figure*}[!t]
  \centering
  \includegraphics[width=0.74\textwidth]{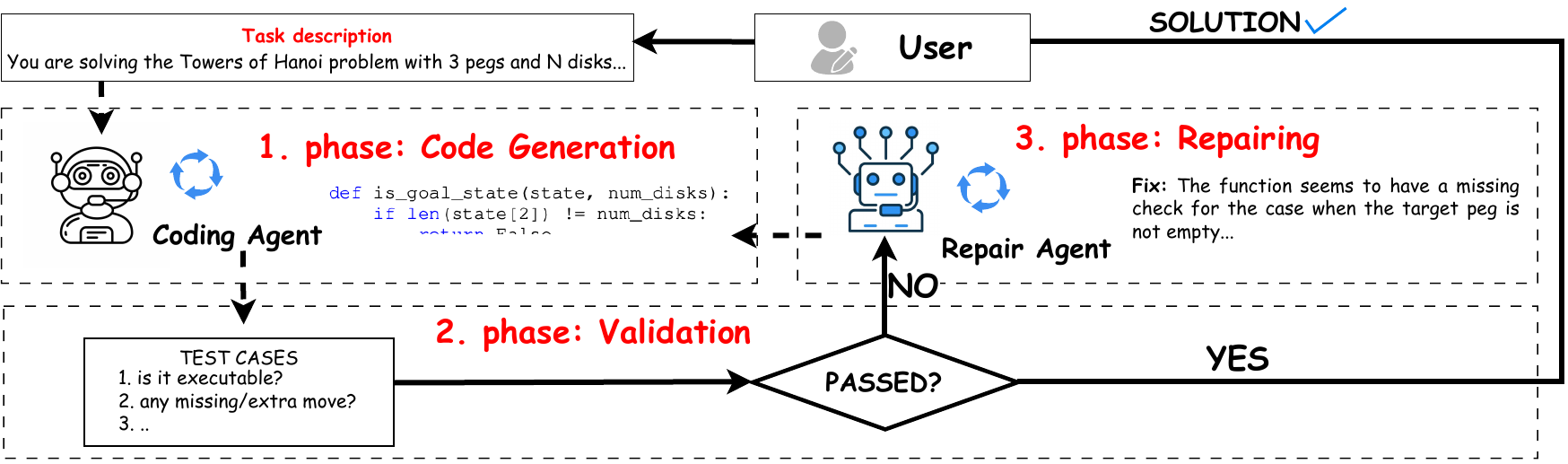}
  \caption{\ac{datos} Architecture Overview} 
  \label{fig:da_tos_loop}
\end{figure*}

\subsection{Performance and Computational Cost}
\label{sec:compcost}
The third dimension, focuses on assessing decision making skill versus computational costs required to provide that level of performance. By performance we refer to AI specific metrics measuring model performance at a single step, such as mean squared error distance or F1 or reward and regret characterizing the entire decision sequence. By computation cost we refer to metrics such as memory footprint, floating point operations, time and energy. This axis goes beyond measuring single domain specific performance metric towards a more holistic cost benefit analysis. This proposal is in line with prior work \cite{wang2025comparing} that has emphasized the importance of understanding computational behavior across planning, reinforcement learning, and LLM-based reasoning, and broader overviews \cite{moerland2022unifying}  of sequence models in sequential decision making that discuss how model architecture and training dynamics impact performance and cost.  In this paper, we focus on time and memory consumption as the primary indicators of computational cost. In addition to measured time and memory, we also provide theoretical memory estimates for each selected method. For reproducibility purposes, we publish the code\footnote{GitHub link to be inserted after double blind review completes.}.

\section{Methods}
\label{sec:methods}

This section presents the selection of diverse AI techniques to be analyzed according to the three-dimensional framework introduced in Section~\ref{sec:3D}. We consider LLM-based agents, a graph-based learning method, and a reinforcement learning–based approach.

\subsection{LLM-based Agents}
\label{sec:LLMAgents}

\subsubsection{\ac{tos}}
\ac{autotos} is a framework that automates search-based planning by using a large language model (LLM) to generate two core components: a successor function (\texttt{succ}), which enumerates valid next states, and a goal predicate (\texttt{isgoal}), which determines whether a state satisfies the task objective. In \ac{toh} domain, \texttt{succ} encodes legal disk moves, while \texttt{isgoal} identifies the configuration in which all disks are stacked on the target peg. The generated functions are embedded in an iterative refinement loop, where a Breadth-First Search (BFS) is used to test correctness and optimality. Detected errors, such as missing transitions or incorrect goal checks, are automatically converted into feedback prompts for the LLM, which regenerates the functions until soundness and completeness are achieved.

\subsubsection{\ac{datos}}
We extend \ac{autotos} by introducing a dual-agent architecture that separates code generation from error interpretation. As illustrated in Figure~\ref{fig:da_tos_loop}, in \ac{datos} \emph{Coding agent} generates implementations of \texttt{succ} and \texttt{isgoal}, while a \emph{Repair agent} analyzes compilation errors and test failures to produce targeted feedback. The system operates in a closed-loop workflow: generated code is validated against predefined test cases, failures are interpreted by the Repair agent, and corrective guidance is fed back to the Coding agent for refinement. This process repeats until all tests pass.

\subsubsection{Estimated Memory}
\label{sec:memory_llm}
The total memory used by the model during inference mainly comes from two parts — 
the model weights and the key--value (KV) cache that stores hidden states for each token in the context. It can be expressed as \( M_{\text{total}} \approx M_{\text{weights}} + M_{\text{KV}} \) where \( M_{\text{weights}} = P \times b \)  depends on the number of model parameters \(P\) 
and the precision \(b\) (bytes per parameter) and $M_{\text{KV}} \approx 2 \times L \times T \times d_{\text{model}} \times b$ is proportional with the number of \(L\), hidden size \(d_{\text{model}}\), 
and sequence length \(T\) (tokens in the conversation).
The factor of 2 accounts for storing both keys and values. 
In practice, \(M_{\text{KV}}\) increases linearly with the token count, while \(M_{\text{weights}}\) remains constant, 
so the total memory grows slowly with longer interactions.

\subsubsection{Experimental Setting for LLM-based Agents}

We use standard (non-reasoning) LLMs instead of explicitly trained reasoning models, including low-parameter variants, to avoid introducing additional inductive biases from specialized reasoning supervision. In our approach, reasoning is externalized through search and verification, ensuring correctness independently of the model’s internal deliberation. This allows performance differences to be attributed to the planning framework rather than hidden reasoning priors, making standard LLMs a more neutral baseline.

In our implementation, we used DeepSeek Coder (deepseek-coder-v2-lite-instruct, 16B), Yi Coder (yi-coder-9b-chat, 9B), and Qwen Coder (qwen2.5-coder-32b-instruct, 32B) as Coding agents, while Gemma (8.5B) and Mistral (7.2B) served as Repair agents. By employing coder fine-tuned LLMs for the Coding agent and general-purpose LLMs for the Repair agent, we seek to improve the effectiveness of smaller-parameter coding models while reducing bias through cross-model validation.

Memory estimation is obtained by combining the model weight with an analytically computed key--value (KV) cache cost, using $M_{\text{KV}}$ from Section \ref{sec:memory_llm}. 
The weight values are taken directly from the quantized GGUF models used in our experiments—DeepSeek (9.65~GB), Yi (4.96~GB), Qwen (11.47~GB), Gemma (4.67~GB), and Mistral (3.83~GB). 
The KV-cache is estimated assuming fp16 precision ($b=2$ bytes), with architectural parameters $L=40, d_{\text{model}}=5120$ for DeepSeek, $L=32, d_{\text{model}}=4096$ for Yi and Mistral, $L=64, d_{\text{model}}=7168$ for Qwen, and $L=28, d_{\text{model}}=3072$ for Gemma, yielding a conservative upper-bound estimate where all tokens are assumed to remain in the context window.

\begin{table*}[tbh]
\caption{Comparison of selected ToH models in terms of their State, Action, Transition, Reward, and Goal Representations.}
\centering
\scriptsize
\setlength{\tabcolsep}{1.5pt}
\renewcommand{\arraystretch}{1.7}
\resizebox{\textwidth}{!}{
\begin{tabular}{|p{1.2cm}|p{3.0cm}|p{2.2cm}|p{4.0cm}|p{2.8cm}|p{2.8cm}|}
    \hline
    \textbf{Model} & \textbf{State (S)} & \textbf{Action (A)} & \textbf{Transition (T)} & \textbf{Reward (R)} & \textbf{Goal (G)} \\
    \hline

    \ac{datos} &
    $\langle$task prompt, code $c$, generated feedback $F(c,e)$ (token sequences) &
    Invoke coding or repair LLM once &
    Probabilistic token generation:
    $P(y_{t+1:t+m}\mid x_{1:t})$ &
     1 = working code, 0 = non-working code &
    Verified \texttt{succ} and \texttt{isgoal} functions \\
    \hline

    \ac{autotos} \cite{katz2024thought} &
    $\langle$task prompt, function, verifier feedback$\rangle$ (token prefixes) &
    Single LLM generation step &
    Probabilistic token generation:
    $P(y_{t+1:t+m}\mid x_{1:t})$ &
    1 = working code, 0 = non-working code &
    Verified \texttt{succ} and \texttt{isgoal} functions \\
    \hline

    Neurosolver \cite{bieszczad2015neurosolver} &
    $s \in \{1,2,3\}^n$ (symbolic disk--peg assignments) &
    Legal disk moves &
    Deterministic:
    $s' = T(s,a)$, guided by goal-conditioned value signal &
    Goal reward; penalties for illegal or redundant moves &
    All disks stacked on target peg (optimal $2^n-1$ steps) \\
    \hline

    FBRL \cite{edwards2018forward} &
    Bit string (disk positions) &
    Valid disk moves (bit changes) &
    Deterministic: $Q(s_t, A_t) \approx \mathbb{E} [R + \gamma \max Q(s_{t+1}, A)]$ &
    +1 valid, -1 invalid, +10 for goal &
    Bit string for disks on Peg C (binary pattern) \\
    \hline
\end{tabular}
}
\label{tab:comparison_mdp_selected}
\end{table*}

\subsection{Neurosolver}
\label{sec:neurosolver}

Neurosolver \cite{bieszczad2015neurosolver} is a graph-based planner that models puzzle state as a node with two divisions. States are encoded as strings, for example in \ac{toh} [123,000,000] for three disks stacked on the first peg. The upper division propagates goal-directed activity backward from the target state, tracing feasible transitions in reverse. The lower division encodes the current configuration and propagates signals forward to execute valid moves. Training builds the state space and transitions, the search phase spreads activation backward from the goal, and the solving phase reconstructs a valid path forward from the start state.

\subsubsection{Estimated Memory} 
\label{sec:neuro_mem}
The memory requirements of Neurosolver can be understood from the way the puzzle’s state space grows as discussed in this section. Because Neurosolver stores both states and transitions (in forward and reverse 
directions), the memory footprint grows in proportion to both of these quantities. If we let $\alpha$ be the average storage cost per state and $\beta$ the average storage cost per edge, the overall memory can be approximated as $M_{neuro}(n) \approx \alpha \cdot 3^n \;+\; \beta \cdot n \cdot 3^n = (\alpha + \beta n)\,3^n$. This formula shows that memory usage increases \emph{exponentially} with the 
number of disks ($3^n$), with an additional \emph{linear factor} in $n$ due to the growing number of possible moves.

\subsubsection{Experimental Setting for Neurosolver}
Neurosolver was implemented such that each disk-number configuration was trained using 100 iterations of random exploration. To ensure robustness, each configuration was trained and solved in ten independent runs, with results averaged across runs reported for metrics such as step count and execution time.

To estimate memory usage, $M_{neuro}$, each state is assigned a constant storage cost of $\alpha = 64$ bytes, corresponding to a compact fixed-size encoding (e.g., $4 \times 4 \times 4$ bytes per peg). Each transition is assigned $\beta = 16$ bytes, representing a stored reference (source–target identifiers). Since both forward and reverse transitions are stored, this cost is doubled using a duplication factor $\kappa = 2$. We further assume an average out-degree of $d = 3$ valid moves per state. These assumptions are combined to estimate the total memory footprint, which is reported in megabytes.

\subsection{Forward-Backward Reinforcement Learning}
\label{sec:fbrl}
\ac{fbrl} \cite{edwards2018forward} is a reinforcement learning method that combines forward and backward exploration during training. Built on the Double Deep Q-Network (DDQN) framework with a replay buffer, FBRL exploits both real and imagined experiences to accelerate learning in sparse-reward environments. In the forward step, the agent updates Q-values using real trajectories collected from the start state, while in the backward step it generates imagined trajectories by predicting reverse transitions from known goal states. Replay buffer is prioritized where transitions with higher temporal-difference errors are sampled more frequently ~\cite{schaul2015prioritized}. 

\subsubsection{Estimated Memory} 
\label{sec:fbrl_memory}
The total memory consumption of FBRL can be decomposed into three principal components: the Q-network, the backward (reverse) model, and the replay buffer. The Q-network learns a mapping \(Q(s,a;\theta)\), and its memory cost is determined by the number of parameters in that network. If the Q-network has \(L_Q\) hidden layers each of width \(h_Q\), with input dimension \(3n\) and output dimension \(\lvert A\rvert\), then its memory cost can be approximated as $M_Q = \alpha_Q \bigl(3n \cdot h_Q + (L_Q - 1)\, h_Q^2 + h_Q \cdot \lvert A\rvert\bigr)$, where \(\alpha_Q\) is the average cost per parameter (in bytes or memory units).

Similarly, the backward model—which predicts reverse transitions or goal-conditioned dynamics—has an analogous architecture. Let it have \(L_B\) hidden layers of width \(h_B\). Then its memory cost is $M_B = \alpha_B \bigl(3n \cdot h_B + (L_B - 1)\, h_B^2 + h_B \cdot \lvert A\rvert\bigr)$ where \(\alpha_B\) is its parameter storage cost.

The replay buffer stores past transitions \((s, a, r, s')\) and thus scales proportionally with the number of stored entries. Denoting the buffer capacity by \(N\) and the average memory per stored tuple by \(\gamma\), we write \(M_R = \gamma N\). Adding up the three components, the total memory consumed by FBRL is $M_{FBRL}(n) \approx (\alpha_Q + \alpha_B)\bigl(3n\,h + (L - 1)\,h^2 + h\,n \bigr) + \gamma\, N$. This expression makes clear that while the network components scale roughly linearly with \(n\) (given fixed architecture), the replay buffer term \(\gamma N\) is independent of \(n\) and may dominate memory usage for sufficiently large \(N\).

\subsubsection{Experimental Setting for FBRL}
We implemented FBRL for \ac{toh} using a string-based state representation. Each disk is encoded according to its peg location, expressed as a binary tuple. For example, the state [(100), (100), (100)] denotes the initial configuration of a three-disk problem, where all disks are placed on the first peg. In the experimental setup, the replay buffer was set to a capacity of 10{,}000 transitions, and the DDQN learning rate was fixed at $5 \times 10^{-4}$.

To estimate the memory consumption of FBRL, we use $M_{FBRL}$ introduced in this section, which decomposes total memory into the Q-Network, Backward Model, and Replay Buffer components. The estimation was based on the architecture parameters: number of disks $n=4$, input dimension $d_{\text{in}} = 3n = 12$, action space $|A| = 3n = 12$, hidden widths $h_Q = 64$ and $h_B = 100$, and two linear layers per network ($L_Q = L_B = 2$). The replay buffer capacity was set to $N = 10{,}000$ transitions, with all values stored in 32-bit floating-point format (4 bytes each).
\section{Analysis from the \ac{mdp}  Perspective}
\label{sec:dim1}
Along the first dimention of the proposed framework in Section \ref{sec:3D}, in this section, we project the methods from Section \ref{sec:methods} to the elements of the \ac{mdp}. The columns represent State, Action, Transition, Reward, and Goal, while each row corresponds to a method, as shown in Table~\ref{tab:comparison_mdp_selected}.

\paragraph*{\ac{datos}}
\begin{figure}[b]
    \centering
    \includegraphics[width=0.79\linewidth]{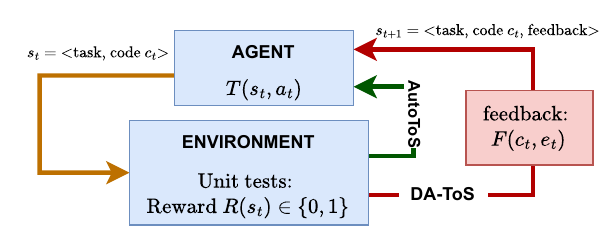}
    \caption{LLM-based Agents Projection to MDP.}
    \label{fig:mdp_form}
\end{figure}
In \ac{datos}, \ac{mdp} is formulated over sequential language-model interactions between a Coding agent and a Repair agent as illustrated in Figure \ref{fig:mdp_form}. Let $t$ denote the iteration step. The initial state is defined as
$
s_0 = \langle \text{task prompt} \rangle,
$
where the task prompt specifies the problem instance (Boxes~1 and~2 from the Annex).
At step $t$, the state is defined as
$
s_t = \langle \text{task prompt},\, c_t,\, F(c_t, e_t) \rangle,
$
where $c_t$ denotes the current candidate code implementation, $e_t$ represents the corresponding evaluation signal (e.g., unit-test or verifier output), and $F(c_t, e_t)$ denotes the feedback generated by the Repair agent based on the code and its evaluation. The action space \(A\) consists of invoking a language model once. An action corresponds either to generating updated code (coding agent) or generating feedback $F(c_t, e_t)$ (repair agent). The transition function is induced by autoregressive generation. 
Let the internal token representation of $s_t$ be denoted by 
$(x_1,\ldots,x_n)$, where $n$ is the current token length. 
The transition probability is defined as:
\begin{equation}
\label{eq:transition_llm}
T(s_t,a_t)=P(y_{n+1:n+m}\mid x_{1:n})
      =\prod_{i=n+1}^{n+m} P(y_i \mid x_{1:n}, y_{<i}),
\end{equation}
where $y_{n+1:n+m}$ corresponds to the newly generated tokens.
The next state is obtained by concatenation,
$
s_{t+1} = s_t \,\Vert\, (y_{n+1},\ldots,y_{n+m}).
$
The reward function is binary, $R(s_t) \in \{0,1\}$, assigning $1$ only when the generated planning functions satisfy soundness and completeness constraints, and $0$ otherwise. Iteration proceeds until $R(s_t)=1$, at which point the goal state is reached. The goal state $G$ is defined as the state in which both \texttt{succ} and \texttt{isgoal} are verified to be sound and complete, ensuring that the \ac{toh} problem instance can be solved correctly with 100\% validity.

\paragraph*{\ac{autotos}}
In \ac{autotos}, the \ac{mdp} follows the same formulation as in \ac{datos} (see Figure~\ref{fig:mdp_form}), but without a separate Repair agent providing feedback. Let $t$ denote the iteration step. The initial state is
$
s_0 = \langle \text{task prompt} \rangle.
$
At step $t$, the state is defined as
$
s_t = \langle \text{task prompt},\, c_t,\, e_t \rangle,
$
where $c_t$ denotes the current candidate implementation (e.g., \texttt{succ}, \texttt{isgoal}), and $e_t$ denotes the verification feedback obtained directly from predefined unit tests and BFS soundness checks. Unlike \ac{datos}, no secondary model generates feedback; the verifier output $e_t$ is directly appended to the state. The action space consists of a single language-model invocation conditioned on $s_t$. The transition dynamics follow the same autoregressive formulation as in Eq.~\ref{eq:transition_llm}, and the next state is obtained via token concatenation,
$
s_{t+1} = s_t \,\Vert\, (y_{n+1},\dots,y_{n+m}).
$
The reward function remains binary, $R(s_t) \in \{0,1\}$, with termination when $R(s_t)=1$. The goal state $G$ is reached when both \texttt{succ} and \texttt{isgoal} are verified to be sound and complete, ensuring correct solution of the \ac{toh} problem.

\paragraph*{Neurosolver}  
In Neurosolver, the problem is modeled as a discrete \ac{mdp}. The state space is defined as
$s_n = \{1,2,3\}^n$, where a state \(s = (s_1,\ldots,s_n)\) encodes the peg position of each disk, and the initial state is $s_0 = \{123,000,000\}$, corresponding to all disks stacked on the source peg. The action space consists of all legal disk moves, and the transition function is deterministic,
$T : s_n \times A \rightarrow s_n, s' = T(s,a)$,
with transitions defined only for legality-preserving actions. Action selection is guided by a goal-conditioned value signal, such that $ \pi(a \mid s, g) \propto \exp\!\big(V_g(s')\big)$, biasing exploration toward states closer to the goal. The reward function assigns maximal reward when the goal is reached, penalizes illegal or redundant moves, and otherwise reflects progress toward the target state. The goal state \(G\) is defined as the configuration where all disks are stacked on the destination peg, yielding the optimal solution length \(2^n - 1\).

\begin{table*}[t]
\scriptsize
\centering
\setlength{\tabcolsep}{3pt}
\renewcommand{\arraystretch}{1.6}
\caption{Human Prior ranking across Architecture, Training, and Inference Stages.}
\label{tab:human_prior}
\begin{tabular}{|p{1.5cm}|p{2.8cm}|c|c|c|c|c|c|c|}
\hline
\textbf{Method} &
\textbf{Model} &
\multicolumn{2}{c|}{\textbf{Architecture}} &
\multicolumn{2}{c|}{\textbf{Training}} &
\multicolumn{3}{c|}{\textbf{Inference}} \\
\cline{3-9}
 &  &
\textbf{Representation} & \textbf{Mechanism} &
\textbf{Knowledge} & \textbf{Learning} &
\textbf{Guidance} & \textbf{Decision} & \textbf{Validation} \\
\hline

\multirow{2}{*}{\ac{datos}}
& DeepSeek + Mistral
& \cellcolor{pastelyellow} & \cellcolor{pastelblue}
& \cellcolor{pastelblue} High & \cellcolor{pastelblue} High
& \cellcolor{pastelyellow} & \cellcolor{pastelgreen} & \cellcolor{pastelyellow} \\
\cline{2-2}\cline{5-6}
& DeepSeek + Gemma
& \multirow{-2}{*}{\cellcolor{pastelyellow} Medium} & \multirow{-2}{*}{\cellcolor{pastelblue} High}
& \cellcolor{pastelyellow} Medium & \cellcolor{pastelyellow} Medium
& \multirow{-2}{*}{\cellcolor{pastelyellow} Medium} & \multirow{-2}{*}{\cellcolor{pastelgreen} Low} & \multirow{-2}{*}{\cellcolor{pastelyellow} Medium} \\
\hline

\multirow{3}{*}{\ac{autotos} \cite{katz2024thought}}
& DeepSeek
& \cellcolor{pastelyellow} & \cellcolor{pastelblue}
& \cellcolor{pastelblue} High & \cellcolor{pastelblue} High
& \cellcolor{pastelblue} & \cellcolor{pastelgreen} & \cellcolor{pastelyellow} \\
\cline{2-2}\cline{5-6}
& Qwen
& \cellcolor{pastelyellow} & \cellcolor{pastelblue}
& \cellcolor{pastelyellow} Medium & \cellcolor{pastelyellow} Medium
& \cellcolor{pastelblue} & \cellcolor{pastelgreen} & \cellcolor{pastelyellow} \\
\cline{2-2}\cline{5-6}
& Yi
& \multirow{-3}{*}{\cellcolor{pastelyellow} Medium} & \multirow{-3}{*}{\cellcolor{pastelblue} High}
& \cellcolor{pastelyellow} Medium & \cellcolor{pastelyellow} Medium
& \multirow{-3}{*}{\cellcolor{pastelblue} High} & \multirow{-3}{*}{\cellcolor{pastelgreen} Low} & \multirow{-3}{*}{\cellcolor{pastelyellow} Medium} \\
\hline

Neurosolver \cite{bieszczad2015neurosolver}
& Graph-based model
& \cellcolor{pastelgreen} Low & \cellcolor{pastelblue} High
& \cellcolor{pastelgreen} Low & \cellcolor{pastelgreen} Low
& \cellcolor{pastelyellow} Medium & \cellcolor{pastelgreen} Low & N/A \\
\hline

FBRL \cite{edwards2018forward}
& Reinforcement Learning
& \cellcolor{pastelgreen} Low & \cellcolor{pastelblue} High
& \cellcolor{pastelblue} High & \cellcolor{pastelyellow} Medium
& \cellcolor{pastelgreen} Low & \cellcolor{pastelgreen} Low & N/A \\
\hline

\end{tabular}
\label{tab:human_ranks}
\end{table*}

\paragraph*{FBRL}  
In FBRL, the state space $s$ is encoded as a bit string of length \(3n\), where each group of three bits specifies the peg location of a disk. For example, with two disks, the initial state is represented as \([1,0,0,1,0,0]\) (both disks on Peg A), and the goal state as \([0,0,1,0,0,1]\) (both disks on Peg C). The action space $A$ consists of legal disk moves, expressed as updates to the relevant 3-bit segment of the bit string while ensuring only valid transitions occur. The transition function $T$ is deterministic; approximated with a DDQN, where transitions are sampled from a replay buffer and updated using the Bellman equation $
Q(s_t, A_t) \approx \mathbb{E} \left[ R + \gamma \max Q(s_{t+1}, A) \right]$. The reward function (R) assigns \(+1\) for valid moves, \(-1\) for illegal moves (e.g., placing a larger disk on top of a smaller one), and a terminal reward of \(+10\) when the goal is reached. The goal state $G$ is the bit-string configuration where all disks are successfully moved to Peg C.

\textbf{Summary:} The projecting of methods to the \ac{mdp} formalism enables in depth analysis and understanding of their details. From the analysis it is possible to notice that the state space for the ToS/LLM-based methods consists of a sequence of tokens and it’s significantly larger than in the case of Neurosolver and FBRL. The state, transition and action structure of the  ToS/LLM rely on probabilistic distributions (input, output tokens and the LLM) while in the case of RL and graph based they rely on graphs encoding states in nodes and transitions on edges that dictate the action. The reward in ToS is binary realized through the verifier loop, while the other two rely on maximizing continuous reward.

\section{Analysis from the Human Prior Perspective}
\label{sec:humanprior}
In this section, we analyze the degree of human prior embedded in each method, as defined in Section~\ref{sec:human_prior}. Table~\ref{tab:human_ranks} reports a stage-wise comparison across Architecture, Training, and Inference, highlighting how human knowledge and design choices shape each system.

\paragraph{Architecture}
In \ac{autotos} and \ac{datos}, representation is implicit and function-based (Table~\ref{tab:architecture}): the model generates successor and goal-checking functions without an explicitly defined symbolic state encoding. This corresponds to a Medium level of human prior, since the task formulation and verification criteria are human-defined, while the internal state representation is learned implicitly by the language model. In contrast, Neurosolver and FBRL use low-level encodings that capture only disk positions on pegs, without higher-level semantics, resulting in a Low level of human prior.

For all methods, the reasoning mechanism, as defined in Table \ref{tab:architecture}, carries a High level of human prior. \ac{autotos} and \ac{datos} rely on Transformer-based language models, whose architectures are fully human-designed. FBRL is built on the Double Deep Q-Network framework, which explicitly specifies the learning and decision-making mechanism. Neurosolver employs a graph-based planning mechanism with manually defined propagation rules. In all cases, the core reasoning process is explicitly designed by humans rather than discovered autonomously.
\paragraph{Training}

For LLM-based methods (\ac{autotos} and \ac{datos}), the level of human prior in knowledge input as defined in Table \ref{tab:training}, depends on the underlying language model. DeepSeek, Yi, and Mistral are trained primarily on large-scale, human-curated datasets, which directly embed substantial human knowledge into the models, resulting in a High level of human prior \cite{guo2024deepseek, young2024yi, chaplot2023albert}. Qwen and Gemma, in contrast, incorporate a significant amount of synthetic data and automated reinforcement learning signals during training, reducing direct human involvement in knowledge specification and leading to a Medium level of human prior \cite{hui2024qwen2, team2024gemma}. FBRL exhibits a High level of human prior, as humans explicitly define the reward structure, penalties, goal conditions, and environment constraints that directly shape the learning signal. In contrast, in Neurosolver, humans do not provide training data, rewards, or explicit transition constraints; the model explores valid moves autonomously based on basic puzzle rules, corresponding to a Low level of human prior.

In \ac{autotos} and \ac{datos}, the learning as defined in Table \ref{tab:training}, process depends on the underlying LLM. DeepSeek is trained primarily through large-scale supervised learning on human-curated code data, resulting in a High level of human prior. Qwen incorporates automated reinforcement learning with synthetic data generation, reducing direct human involvement and leading to a Medium level of human prior \cite{hui2024qwen2}. Yi relies on continued pretraining from an existing base model using curated data \cite{young2024yi}, which also corresponds to a Medium level. Among repair agents, Mistral is trained via supervised instruction tuning with human-curated datasets, yielding High human prior, whereas Gemma explicitly employs reinforcement learning from human feedback \cite{team2024gemma} (RLHF), introducing partial automation and resulting in a Medium level. In contrast, Neurosolver and FBRL do not use human feedback–based learning. Neurosolver learns autonomously through exploration over valid state transitions, and FBRL updates its policy via predefined reinforcement learning rules. Since no human feedback is involved during learning, both correspond to a Low level of human prior.

\paragraph{Inference}
Guidance, as defined in Table \ref{tab:inference}, differs across methods in how task constraints are provided at inference time. \ac{autotos} exhibits a High level of guidance, as human-provided test cases and error types are directly used to steer the refinement loop during inference. In \ac{datos}, this guidance is partially automated: although human-defined test cases remain, error interpretation and correction are handled by a secondary language model, resulting in a Medium level of guidance. Neurosolver also exhibits Medium guidance, since human-defined legality rules constrain admissible moves during execution without interactive input. In contrast, FBRL operates with Low guidance, executing a learned policy without task-specific input or constraints at inference time.

For all evaluated methods, the level of human prior in decision-making as defined in Table \ref{tab:inference}, is Low. Action selection and planning are performed by the systems themselves during inference, without direct human involvement across \ac{autotos}, \ac{datos}, Neurosolver, and FBRL.

Validation, as defined in Table \ref{tab:inference}, differs substantially across methods. In \ac{autotos} and \ac{datos}, humans explicitly define test cases that are used to verify solution correctness, resulting in a Medium level of human prior in validation. In contrast, Neurosolver and FBRL do not employ an explicit validation stage: correctness is determined implicitly by goal satisfaction within the environment, making validation not applicable for these methods.

\textbf{Summary:} The human prior ranking of the methods makes a significant view on understanding autonomous level. In general, it can be clearly seen that ToS/LLM-based approaches need more human insight compare to FBRL/Neurosolver. Due to constrain-based architectural design, FBRL and Neurosolver don’t need verification after final decision, this shows clear superiority compare to ToS/LLM-based methods. While all four paradigm include high human prior in their architectural design, FBRL and Neurosolver don’t need verification after final decision, due to stricter constraining at architectural stage.
\begin{table*}[t!]
\centering
\scriptsize
\setlength{\tabcolsep}{3pt}
\renewcommand{\arraystretch}{1.5}

\begin{threeparttable}
\caption{Comprehensive Performance and Memory Comparison across \ac{datos}, \ac{autotos}, FBRL, and Neurosolver.
All values are reported as mean $\pm$ MAD$^{*}$.}
\label{tab:results_merged}

\begin{tabular}{|l|c|c|c|r|c|c|}
\hline
\textbf{Model Setup} & \textbf{Disk} & \textbf{Att. / Steps} & \textbf{Tokens} & \textbf{Time (s)} & \textbf{Total E.Mem (MB)} & \textbf{Total O.Mem (MB)} \\ \hline

\rowcolor{gray!10}
\multicolumn{7}{|l|}{\textbf{\ac{datos}}} \\ \hline
DeepSeek + Gemma 
 & $n$
 & $6.8 \pm 1.5$
 & $17998.2 \pm 3552.5$
 & $209.21 \pm 47.84$
 & $1.82 \times 10^{4}$
 & $1.55 \times 10^{4} \pm 0$ \\
DeepSeek + Mistral
 & $n$
 & $5.0 \pm 1.5$
 & $6718.4 \pm 1571.0$
 & $32.69 \pm 10.37$
 & $1.55 \times 10^{4}$
 & $1.43 \times 10^{4} \pm 0$ \\ \hline

\rowcolor{gray!10}
\multicolumn{7}{|l|}{\textbf{\ac{autotos} \cite{katz2024thought}}} \\ \hline
DeepSeek
 & $n$
 & $4.2 \pm 1.0$
 & $5257.5 \pm 975.0$
 & $23.61 \pm 8.80$
 & $1.40 \times 10^{4}$
 & $1.16 \times 10^{4} \pm 0$ \\
Yi
 & $n$
 & $2.0 \pm 0.0$
 & $1162.3 \pm 64.5$
 & $26.37 \pm 5.06$
 & $5.47 \times 10^{3}$
 & $6.16 \times 10^{3} \pm 0$ \\
Qwen
 & $n$
 & $2.0 \pm 0.0$
 & $831.4 \pm 3.5$
 & $33.56 \pm 1.16$
 & $1.27 \times 10^{4}$
 & $1.23 \times 10^{4} \pm 0$ \\ \hline

\rowcolor{gray!10}
\multicolumn{7}{|l|}{\textbf{Neurosolver \cite{bieszczad2015neurosolver}}} \\ \hline
\multirow{6}{*}{} 
Graph-based & 2 & 3   & --- & $12.06 \pm 0.04$
 & $1.37 \times 10^{-3}$
 & $5.41 \times 10^{-3} \pm 9.77 \times 10^{-7}$ \\
           & 3 & 7   & --- & $19.75 \pm 0.18$
 & $4.12 \times 10^{-3}$
 & $5.62 \times 10^{-3} \pm 8.30 \times 10^{-6}$ \\
           & 4 & 15  & --- & $28.13 \pm 0.17$
 & $1.24 \times 10^{-2}$
 & $1.40 \times 10^{-2} \pm 0$ \\
           & 5 & 31  & --- & $38.00 \pm 0.35$
 & $3.71 \times 10^{-2}$
 & $2.51 \times 10^{-2} \pm 0$ \\
           & 6 & 63  & --- & $49.76 \pm 0.45$
 & $1.11 \times 10^{-1}$
 & $1.05 \times 10^{-1} \pm 0$ \\
           & 7 & 127 & --- & $60.40 \pm 0.15$
 & $3.34 \times 10^{-1}$
 & $3.30 \times 10^{-1} \pm 5.47 \times 10^{--5}$ \\ \hline

\rowcolor{gray!10}
\multicolumn{7}{|l|}{\textbf{FBRL \cite{edwards2018forward}}} \\ \hline
\multirow{3}{*}{Reinforcement Learning}
 & 2 & --- & --- & $2.42 \pm 0.07$
 & \makecell{$6.65 \times 10^{-1}$ \\ (0.018 + 0.037 + 0.61)}
 & \makecell{$3.29 \times 10^{-1} \pm 6.25 \times 10^{-3}$ \\ (0.004 + 0.005 + 0.32)} \\
 & 3 & --- & --- & $7.68 \pm 0.27$
 & \makecell{$9.03 \times 10^{-1}$ \\ (0.020 + 0.043 + 0.84)}
 & \makecell{$7.03 \times 10^{-1} \pm 0$ \\ (0.005 + 0.008 + 0.69)} \\
 & 4 & --- & --- & $31.20 \pm 2.02$
 & \makecell{$1.14 \times 10^{0}$ \\ (0.022 + 0.048 + 1.07)}
 & \makecell{$9.37 \times 10^{-1} \pm 0$ \\ (0.007 + 0.010 + 0.92)} \\ \hline

\end{tabular}
\begin{tablenotes}
\scriptsize
\item[$^{*}$] MAD denotes the Median Absolute Deviation, defined as 
$\mathrm{MAD} = \mathrm{median}\!\left(\lvert x_i - \mathrm{median}(x) \rvert\right)$,
computed over repeated runs.

\end{tablenotes}
\end{threeparttable}

\end{table*}
\section{Analysis from the Skill and Computational Cost Perspective}
\label{sec:results}

In this section, we analyze the skill and computational cost of each method, as defined in Section~\ref{sec:compcost} by focusing on the cost to solve the puzzle correctly. We consider the methods and experimental settings outlined in Section~\ref{sec:methods}.  In the analysis on \ac{toh} problem depicted in Figure~\ref{fig:TOH} we  assume three pegs and an number of disks $n\in {2, 3, ...7}$. Table~\ref{tab:results_merged} presents a comprehensive comparison of experimental results across \ac{datos}, \ac{autotos}, FBRL, and Neurosolver, with all values reported as mean ± Median Absolute Deviation. The table reports the model configuration, including the LLM combinations used for LLM-based agents, the number of disks, the number of attempts or steps, token usage for LLM-based methods, total execution time (including both training and solving), and both estimated and observed memory consumption.

\subsection{LLM-based Agents}
For LLM-based approaches, the reported results correspond to the combined generation of the Successor and Goal functions. As a result, the number of attempts (LLM call) reflects successful generation for both components, with one step required for the Goal function and one for the Successor function in the optimal case.
\paragraph*{\ac{datos}}
For \ac{datos}, it is evident that the double-agent design incurs higher computational cost in terms of both token usage and execution time. Although the DeepSeek+Gemma and DeepSeek+Mistral configurations exhibit comparable memory consumption, the Mistral-based setup is markedly more efficient, requiring fewer steps, fewer tokens, and shorter execution time, while also exhibiting lower variance across runs. Specifically, the DeepSeek+Gemma configuration consumes over 17k tokens and approximately 209 seconds, whereas DeepSeek+Mistral completes the task with roughly 6.5k tokens in about 32 seconds. The estimated memory consumption exceeds the observed usage, as the estimation assumes simultaneous high-precision execution of both agents. In practice, quantized deployments such as DeepSeek+Gemma (17.73~GB $\rightarrow$ 15.13~GB) and DeepSeek+Mistral (15.12~GB $\rightarrow$ 14.00~GB) use compressed weights and lighter key--value caches, resulting in a lower actual memory footprint.

\paragraph*{Auto-ToS}
For \ac{autotos}, the results indicate a substantially lower computational overhead compared to \ac{datos}, as the approach relies on a single LLM without an explicit repair agent. Across all configurations, \ac{autotos} typically converges in one to three attempts, with lower token usage and shorter execution times. In particular, Qwen and Yi achieve the most efficient performance, requiring fewer than 1k tokens on average, completing execution in approximately 25–35 seconds, and consistently producing the optimal step size, whereas DeepSeek exhibits higher token usage and additional refinement cycles in the Successor function, leading to increased attempts and longer execution times. In terms of memory consumption, \ac{autotos} follows the expected trend based on model size, with DeepSeek exhibiting the largest footprint ($\approx$11.3~GB observed), followed by Qwen ($\approx$12.0~GB) and Yi ($\approx$6.0~GB). As with \ac{datos}, estimated memory exceeds observed usage due to conservative precision and cache assumptions, while quantized execution reduces the actual footprint during inference.

\subsection{Neurosolver}
Across all configurations, Neurosolver consistently achieves the optimal number of steps ($2^n - 1$), while runtime and memory consumption increase with puzzle size. Runtime grows gradually from 12.06 seconds for two disks to 60.40 seconds for seven disks, remaining below one minute even for the largest instance. In contrast, memory consumption increases more sharply, from 0.005~MB to 0.321~MB, reflecting the exponential growth of the state space. While observed memory usage remains within the same order of magnitude as theoretical estimates, deviations are expected due to random exploration and higher-level state representations in the implementation. Overall, this behavior aligns with the theoretical scaling law from Section \ref{sec:neuro_mem}, highlighting the trade-off between guaranteed optimality and rapidly growing memory requirements.

\subsection{Forward-backward Reinforcement Learning}

For FBRL, computational cost increases rapidly with the number of disks, primarily due to its dual exploration mechanism that combines forward trajectories from the initial state with backward rollouts from the goal. Runtime grows from approximately 2.4 seconds for two disks to over 31 seconds for four disks, reflecting the increased training effort required as the search space expands. In terms of memory consumption, the breakdown shown in the table highlights contributions from the Q-Network, Backward Model, and Replay Buffer. While the Q-Network and Backward Model remain lightweight and scale linearly with the input and output dimensions of $3n$, their memory usage is negligible compared to the replay buffer. In contrast, experience storage dominates the overall footprint, increasing from about 0.32~MB for two disks to approximately 0.92~MB for four disks. Both observed and estimated results indicate that FBRL’s memory behavior is driven primarily by replay buffer growth rather than network complexity, resulting in a predictable but increasingly costly memory profile.

\textbf{Summary:} The skill to solve the ToH puzzle and computational comparison illustrates quantitative insight about pros and cons in terms of memory and time. From the table, it is clear that ToS/LLM-based approaches need large amount of memory resource compared to Neurosolver and FBRL. On the other hand  the ToS/LLM-based ones are not limited by about disk numbers. Focusing on FBRL vs Neurosolver, it seems that Neurosolver has superior scalability performance since time and memory does not increase as much as for FBRL.

% Example mock-up statements suitable for this section:  

% The "Human Prior" analysis suggest that we should stop training end-to-end RL for logic puzzles?

% Through the 3D analysis, we demonstrate that hybrid agentic workflows (DA-ToS) represent the next phase of CI by balancing human prior with computational autonomy.

% Visuals: Create a "Radar Chart" (Spider Plot) for each method (Neurosolver vs. FBRL vs. DA-ToS) along your three axes. This visual comparison is very "Transactions-style" and helps the reader grasp the trade-offs instantly. Example Fig 9 here: https://ieeexplore.ieee.org/document/11311988

\section{Conclusion} \label{sec:con}
In this paper we addressed the challenge that existing evaluations of sequential decision making methods largely rely on isolated pros–cons analyses, which fail to reveal how design choices redistribute complexity across system components. To address this gap, we proposed a three-dimensional characterization that jointly captures \ac{mdp} formulation, human prior, and skill–cost trade-offs. Using the proposed  framework, three representative graph-based, reinforcement learning, and large language model (LLM)-based approaches and the Tower of Hanoi puzzle we found that in LLM-based methods, where the action space is weakly constrained by design, : 1) correctness is enforced through increased human insight at inference time, primarily via verification and correction loops and 2) complexity is shifted  to computation, resulting in significantly higher memory and runtime costs compared to graph-based and reinforcement learning models. Our findings enable structured cross-paradigm comparison linking theoretical formulation with empirical performance. The framework provides a principled basis for analyzing such trade-offs across AI techniques even  beyond the \ac{toh}.

% \section*{Acknowledgment}

% This work was supported by the Slovenian Research Agency (P2-0016).



\bibliographystyle{IEEEtran}
\bibliography{references}

@article{schaul2015prioritized,
  title={Prioritized experience replay},
  author={Schaul, Tom and Quan, John and Antonoglou, Ioannis and Silver, David},
  journal={arXiv preprint arXiv:1511.05952},
  year={2015}
}

@article{katz2024thought,
  title={Thought of search: Planning with language models through the lens of efficiency},
  author={Katz, Michael and Kokel, Harsha and Srinivas, Kavitha and Sohrabi Araghi, Shirin},
  journal={Advances in Neural Information Processing Systems},
  volume={37},
  pages={138491--138568},
  year={2024}
}

@article{estermann2024puzzles,
  title={Puzzles: A benchmark for neural algorithmic reasoning},
  author={Estermann, Benjamin and Lanzend{\"o}rfer, Luca and Niedermayr, Yannick and Wattenhofer, Roger},
  journal={Advances in Neural Information Processing Systems},
  volume={37},
  pages={127059--127098},
  year={2024}
}

@article{hawkins2025enigme,
  title={Enigme: Generative Text Puzzles for Evaluating Reasoning in Language Models},
  author={Hawkins, John},
  journal={arXiv preprint arXiv:2505.04914},
  year={2025}
}

@article{merrotsy2015tower,
  title={The Tower of Hanoi and inductive logic},
  author={Merrotsy, Peter},
  journal={Australian Senior Mathematics Journal},
  volume={29},
  number={1},
  pages={16--24},
  year={2015}
}

@article{anderson1992induction,
  title={Induction, recursion, and the Towers of Hanoi},
  author={Anderson, Oliver D},
  journal={International Journal of Mathematical Education in Science and Technology},
  volume={23},
  number={3},
  pages={339--343},
  year={1992},
  publisher={Taylor \& Francis}
}

@article{shojaee2025illusion,
  title={The illusion of thinking: Understanding the strengths and limitations of reasoning models via the lens of problem complexity},
  author={Shojaee, Parshin and Mirzadeh, Iman and Alizadeh, Keivan and Horton, Maxwell and Bengio, Samy and Farajtabar, Mehrdad},
  journal={arXiv preprint arXiv:2506.06941},
  year={2025}
}

@article{mehiri2024weighted,
  title={The weighted Tower of Hanoi},
  author={Mehiri, El-Mehdi and Belbachir, Hacene},
  journal={Discrete Mathematics, Algorithms and Applications},
  volume={16},
  number={05},
  pages={2350051},
  year={2024},
  publisher={World Scientific}
}

@incollection{aguilar2016application,
  title={The application of Hanoi towers game in logistics management},
  author={Aguilar, Jos{\'e} Alberto Hern{\'a}ndez and Mayo, Augusto Renato P{\'e}rez and Ortu{\~n}o, Santiago Yip and Ochoa-Zezzatti, Alberto and Gallegos, Julio C{\'e}sar Ponce},
  booktitle={Handbook of Research on Military, Aeronautical, and Maritime Logistics and Operations},
  pages={422--433},
  year={2016},
  publisher={IGI Global Scientific Publishing}
}

@inproceedings{havur2013case,
  title={A case study on the tower of hanoi challenge: Representation, reasoning and execution},
  author={Havur, Giray and Haspalamutgil, Kadir and Palaz, Can and Erdem, Esra and Patoglu, Volkan},
  booktitle={2013 IEEE International Conference on Robotics and Automation},
  pages={4552--4559},
  year={2013},
  organization={IEEE}
}

@article{edwards2018forward,
  title={Forward-backward reinforcement learning},
  author={Edwards, Ashley D and Downs, Laura and Davidson, James C},
  journal={arXiv preprint arXiv:1803.10227},
  year={2018}
}

@inproceedings{bieszczad2015neurosolver,
  title={Neurosolver learning to solve Towers of Hanoi puzzles},
  author={Bieszczad, Andrzej and Kuchar, Skyler},
  booktitle={2015 7th International Joint Conference on Computational Intelligence (IJCCI)},
  volume={3},
  pages={28--38},
  year={2015},
  organization={IEEE}
}

@inproceedings{stewart2011neural,
  title={Neural cognitive modelling: A biologically constrained spiking neuron model of the Tower of Hanoi task},
  author={Stewart, Terrence and Eliasmith, Chris},
  booktitle={Proceedings of the Annual Meeting of the Cognitive Science Society},
  volume={33},
  number={33},
  year={2011}
}

@inproceedings{agravante2023learning,
  title={Learning Neuro-Symbolic World Models with Logical Neural Networks},
  author={Agravante, Don Joven and Kimura, Daiki and Tatsubori, Michiaki},
  booktitle={PRL Workshop Series Bridging the Gap Between AI Planning and Reinforcement Learning},
  year={2023}
}

@article{li2025puzzleworld,
  title={PuzzleWorld: A Benchmark for Multimodal, Open-Ended Reasoning in Puzzlehunts},
  author={Li, Hengzhi and Jiang, Brendon and Naehu, Alexander and Song, Regan and Zhang, Justin and Tjandrasuwita, Megan and Ekbote, Chanakya and Chen, Steven-Shine and Balachandran, Adithya and Dai, Wei and others},
  journal={arXiv preprint arXiv:2506.06211},
  year={2025}
}

@article{chen2024ai,
  title={AI Planning: A Primer and Survey (Preliminary Report)},
  author={Chen, Dillon Z and Verma, Pulkit and Srivastava, Siddharth and Katz, Michael and Thi{\'e}baux, Sylvie},
  journal={arXiv preprint arXiv:2412.05528},
  year={2024}
}

@article{guo2024deepseek,
  title={DeepSeek-Coder: When the Large Language Model Meets Programming--The Rise of Code Intelligence},
  author={Guo, Daya and Zhu, Qihao and Yang, Dejian and Xie, Zhenda and Dong, Kai and Zhang, Wentao and Chen, Guanting and Bi, Xiao and Li, YK and others},
  journal={arXiv preprint arXiv:2401.14196},
  year={2024}
}

@article{young2024yi,
  title={Yi: Open foundation models by 01. ai},
  author={Young, Alex and Chen, Bei and Li, Chao and Huang, Chengen and Zhang, Ge and Zhang, Guanwei and Wang, Guoyin and Li, Heng and Zhu, Jiangcheng and Chen, Jianqun and others},
  journal={arXiv preprint arXiv:2403.04652},
  year={2024}
}

@article{hui2024qwen2,
  title={Qwen2. 5-coder technical report},
  author={Hui, Binyuan and Yang, Jian and Cui, Zeyu and Yang, Jiaxi and Liu, Dayiheng and Zhang, Lei and Liu, Tianyu and Zhang, Jiajun and Yu, Bowen and Lu, Keming and others},
  journal={arXiv preprint arXiv:2409.12186},
  year={2024}
}

@article{vaswani2017attention,
  title={Attention is all you need},
  author={Vaswani, Ashish and Shazeer, Noam and Parmar, Niki and Uszkoreit, Jakob and Jones, Llion and Gomez, Aidan N and Kaiser, {\L}ukasz and Polosukhin, Illia},
  journal={Advances in neural information processing systems},
  volume={30},
  year={2017}
}

@article{team2024gemma,
  title={Gemma: Open models based on gemini research and technology},
  author={Team, Gemma and Mesnard, Thomas and Hardin, Cassidy and Dadashi, Robert and Bhupatiraju, Surya and Pathak, Shreya and Sifre, Laurent and Rivi{\`e}re, Morgane and Kale, Mihir Sanjay and Love, Juliette and others},
  journal={arXiv preprint arXiv:2403.08295},
  year={2024}
}

@article{chaplot2023albert,
  title={Mistral 7B},
  author = {Albert Q. Jiang and Alexandre Sablayrolles and Arthur Mensch and Chris Bamford and Devendra Singh Chaplot and Diego de las Casas and Florian Bressand and Gianna Lengyel and Guillaume Lample and Lucile Saulnier and L{\'e}lio Renard Lavaud and Marie-Anne Lachaux and Pierre Stock and Teven Le Scao and Thibaut Lavril and Thomas Wang and Timoth{\'e}e Lacroix and William El Sayed},
  journal={arXiv preprint arXiv:2310.06825},
  year={2023}
}

@inproceedings{bai2020provable,
  title={Provable self-play algorithms for competitive reinforcement learning},
  author={Bai, Yu and Jin, Chi},
  booktitle={International conference on machine learning},
  pages={551--560},
  year={2020},
  organization={PMLR}
}

@inproceedings{silver2022pddl,
  title={PDDL planning with pretrained large language models},
  author={Silver, Tom and Hariprasad, Varun and Shuttleworth, Reece S and Kumar, Nishanth and Lozano-P{\'e}rez, Tom{\'a}s and Kaelbling, Leslie Pack},
  booktitle={NeurIPS 2022 foundation models for decision making workshop},
  year={2022}
}

@article{tamar2016value,
  title={Value iteration networks},
  author={Tamar, Aviv and Wu, Yi and Thomas, Garrett and Levine, Sergey and Abbeel, Pieter},
  journal={Advances in neural information processing systems},
  volume={29},
  year={2016}
}

@inproceedings{alshahwan2024automated,
  title={Automated unit test improvement using large language models at meta},
  author={Alshahwan, Nadia and Chheda, Jubin and Finogenova, Anastasia and Gokkaya, Beliz and Harman, Mark and Harper, Inna and Marginean, Alexandru and Sengupta, Shubho and Wang, Eddy},
  booktitle={Companion Proceedings of the 32nd ACM International Conference on the Foundations of Software Engineering},
  pages={185--196},
  year={2024}
}

@article{welleckgenerating,
  title={Generating sequences by learning to self-correct},
  author={Welleck, Sean and Lu, Ximing and West, Peter and Brahman, Faeze and Shen, Tianxiao and Khashabi, Daniel and Choi, Yejin},
  journal={arXiv preprint arXiv:2211.00053},
  year={2022}
}

@inproceedings{wachi2020safe,
  title={Safe reinforcement learning in constrained markov decision processes},
  author={Wachi, Akifumi and Sui, Yanan},
  booktitle={International Conference on Machine Learning},
  pages={9797--9806},
  year={2020},
  organization={PMLR}
}

@article{wang2023describe,
  title={Describe, explain, plan and select: interactive planning with llms enables open-world multi-task agents},
  author={Wang, Zihao and Cai, Shaofei and Chen, Guanzhou and Liu, Anji and Ma, Xiaojian Shawn and Liang, Yitao},
  journal={Advances in Neural Information Processing Systems},
  volume={36},
  pages={34153--34189},
  year={2023}
}

@article{zhou2022least,
  title={Least-to-most prompting enables complex reasoning in large language models},
  author={Zhou, Denny and Sch{\"a}rli, Nathanael and Hou, Le and Wei, Jason and Scales, Nathan and Wang, Xuezhi and Schuurmans, Dale and Cui, Claire and Bousquet, Olivier and Le, Quoc and others},
  journal={arXiv preprint arXiv:2205.10625},
  year={2022}
}

@inproceedings{van2016deep,
  title={Deep reinforcement learning with double q-learning},
  author={Van Hasselt, Hado and Guez, Arthur and Silver, David},
  booktitle={Proceedings of the AAAI conference on artificial intelligence},
  volume={30},
  number={1},
  year={2016}
}

@inproceedings{hegde2024self,
  title={Self supervised LLM customizer (SSLC): Customizing LLMs on unlabeled data to enhance contextual question answering},
  author={Hegde, Raveendra R and Sharma, Saurabh},
  booktitle={Proceedings of the 4th International Conference on AI-ML Systems},
  pages={1--11},
  year={2024}
}

@article{moerland2022unifying,
  title={A unifying framework for reinforcement learning and planning},
  author={Moerland, Thomas M and Broekens, Joost and Plaat, Aske and Jonker, Catholijn M},
  journal={Frontiers in Artificial Intelligence},
  volume={5},
  pages={908353},
  year={2022},
  publisher={Frontiers Media SA}
}

@article{qin2024uno,
  title={UNO arena for evaluating sequential decision-making capability of large language models},
  author={Qin, Zhanyue and Wang, Haochuan and Liu, Deyuan and Song, Ziyang and Fan, Cunhang and Lv, Zhao and Wu, Jinlin and Lei, Zhen and Tu, Zhiying and Chu, Dianhui and others},
  journal={arXiv preprint arXiv:2406.16382},
  year={2024}
}

@article{burden2025paradigms,
  title={Paradigms of AI evaluation: Mapping goals, methodologies and culture},
  author={Burden, John and Te{\v{s}}i{\'c}, Marko and Pacchiardi, Lorenzo and Hern{\'a}ndez-Orallo, Jos{\'e}},
  journal={arXiv preprint arXiv:2502.15620},
  year={2025}
}

@article{wang2025comparing,
  title={Comparing AI and human decision-making mechanisms in daily collaborative experiments},
  author={Wang, Linghao and Jiang, Zheyuan and Hu, Chenke and Zhao, Jun and Zhu, Zheng and Chen, Xiqun and Wang, Ziyi and Liu, Tianming and He, Guibing and Yin, Yafeng and others},
  journal={iScience},
  volume={28},
  number={6},
  year={2025},
  publisher={Elsevier}
}

@article{murugesan2025rise,
  title={The rise of agentic AI: implications, concerns, and the path forward},
  author={Murugesan, San},
  journal={IEEE Intelligent Systems},
  volume={40},
  number={2},
  pages={8--14},
  year={2025},
  publisher={IEEE}
}

@article{castelfranchi1998modelling,
  title={Modelling social action for AI agents},
  author={Castelfranchi, Cristiano},
  journal={Artificial intelligence},
  volume={103},
  number={1-2},
  pages={157--182},
  year={1998},
  publisher={Elsevier}
}

@article{brown2020language,
  title={Language models are few-shot learners},
  author={Brown, Tom and Mann, Benjamin and Ryder, Nick and Subbiah, Melanie and Kaplan, Jared D and Dhariwal, Prafulla and Neelakantan, Arvind and Shyam, Pranav and Sastry, Girish and Askell, Amanda and others},
  journal={Advances in neural information processing systems},
  volume={33},
  pages={1877--1901},
  year={2020}
}

@inproceedings{
wang2025generalization,
title={Generalization v.s. Memorization: Tracing Language Models{\textquoteright} Capabilities Back to Pretraining Data},
author={Xinyi Wang and Antonis Antoniades and Yanai Elazar and Alfonso Amayuelas and Alon Albalak and Kexun Zhang and William Yang Wang},
booktitle={The Thirteenth International Conference on Learning Representations},
year={2025},
url={https://openreview.net/forum?id=IQxBDLmVpT}
}

@article{guo2025deepseek,
  title={Deepseek-r1: Incentivizing reasoning capability in llms via reinforcement learning},
  author={Guo, Daya and Yang, Dejian and Zhang, Haowei and Song, Junxiao and Zhang, Ruoyu and Xu, Runxin and Zhu, Qihao and Ma, Shirong and Wang, Peiyi and Bi, Xiao and others},
  journal={arXiv preprint arXiv:2501.12948},
  year={2025}
}

@article{wei2022chain,
  title={Chain-of-thought prompting elicits reasoning in large language models},
  author={Wei, Jason and Wang, Xuezhi and Schuurmans, Dale and Bosma, Maarten and Xia, Fei and Chi, Ed and Le, Quoc V and Zhou, Denny and others},
  journal={Advances in neural information processing systems},
  volume={35},
  pages={24824--24837},
  year={2022}
}
%

\end{document}